\documentclass{article}
\usepackage{spconf,amsmath,epsfig}

\let\OLDthebibliography\thebibliography
\renewcommand\thebibliography[1]{
  \OLDthebibliography{#1}
  \setlength{\parskip}{0pt}
  \setlength{\itemsep}{0pt plus 0.3ex}
}

\usepackage{marvosym}
\usepackage{amsmath,amsfonts}
\usepackage{algorithmic}
\usepackage{array}
\usepackage{textcomp}
\usepackage{stfloats}
\usepackage{url}
\usepackage{verbatim}
\usepackage{graphicx}
\usepackage{cite}
\usepackage{multirow}
\usepackage[table,xcdraw]{xcolor}

\usepackage{hyperref}
\usepackage[capitalize]{cleveref}

\Crefname{section}{Section}{Sections}
\Crefname{table}{Table}{Tables}
\Crefname{figure}{Figure}{Figures}
\begin{document}\sloppy

\def\x{{\mathbf x}}
\def\L{{\cal L}}

\title{Exploiting Self-Supervised Constraints in Image Super-Resolution}
%
\name{Gang Wu, Junjun Jiang\textsuperscript{\Letter}
\thanks{Corresponding author: Junjun Jiang (jiangjunjun@hit.edu.cn).}, 
Kui Jiang, and Xianming Liu}

\address{Faculty of Computing, Harbin Institute of Technology, Harbin 150001, China}

\maketitle

\begin{abstract}
Recent advances in self-supervised learning, predominantly studied in high-level visual tasks, have been explored in low-level image processing. This paper introduces a novel self-supervised constraint for single image super-resolution, termed SSC-SR. SSC-SR uniquely addresses the divergence in image complexity by employing a dual asymmetric paradigm and a target model updated via exponential moving average to enhance stability. The proposed SSC-SR framework works as a plug-and-play paradigm and can be easily applied to existing SR models. Empirical evaluations reveal that our SSC-SR framework delivers substantial enhancements on a variety of benchmark datasets, achieving an average increase of 0.1 dB over EDSR and 0.06 dB over SwinIR. In addition, extensive ablation studies corroborate the effectiveness of each constituent in our SSC-SR framework. Codes are available at \url{https://github.com/Aitical/SSCSR}.
\end{abstract}
\begin{keywords}
Image Super-Resolution, Contrastive Learning, Self-Supervised Learning
\end{keywords}

\section{Introduction}
Single Image Super-Resolution (SISR) is a fundamental task in image processing that strives to reconstruct high-resolution images from their low-resolution counterparts \cite{survey_sisr_deep21,ACMComputingSurvey}. As an inherently ill-posed problem, SISR has undergone significant evolution with the advent of deep learning, shifting from early convolutional neural network (CNN) methodologies to intricate transformer-based designs \cite{EDSR,RCAN,NLSN,SwinIR,HAT}. Commonly, enhancements in SISR performance have been realized through the exploration of more effective architectural designs. Despite these advances, the challenge of accurately mapping low-resolution to high-resolution images persists.

In parallel, self-supervised learning has garnered considerable interest, demonstrating impressive results in high-level tasks \cite{contrastive_survey}. Within this domain, several methods have been introduced that incorporate contrastive learning into low-level tasks, offering substantive attempts and promising directions \cite{contrastive_dehazing,PCL,MCIR}. These approaches commonly leverage informative negative samples to augment the standard reconstruction loss with contrastive regularization, which can lead to more robust solutions or enhance the fidelity of results. This leads to an intriguing inquiry: \textit{Can self-supervised information be leveraged to further advance current SISR techniques?}

In contrast to existing methods, we revisit the vanilla learning process of the SISR model and introducing a novel self-supervised constraint paradigm for image super-resolution (SSC-SR). Before we dive into our method, we first examine the learning process in current SR networks. It is observed that while smooth areas in images are easily super-resolved, complex regions with rich edges or textures, due to the ill-posed nature of SISR, pose greater challenges. Our approach focus on this issue by using the divergence between current and previous model outputs as a self-supervised constraint. This method not only stabilizes the representation of smooth areas but also emphasizes uncertain regions.

Drawing on these insights, we introduce the Self-Supervised Constraint for Super-Resolution (SSC-SR), a method that leverages self-supervision to refine and stabilize super-resolution (SR) techniques. The cornerstone of SSC-SR is a dual asymmetric framework that unites an online SR model with a target model, the latter continually refined via an exponential moving average (EMA). Additionally, a projection head is employed to facilitate the self-supervised constraint. Our comprehensive experiments substantiate SSC-SR's superior performance, showcasing its effectiveness and versatility across diverse datasets. 

Within this study, we critically assess the end-to-end training pipeline for SISR tasks and introduce a pioneering self-supervised constraint paradigm. This paradigm is versatile and efficacious, enhancing the performance of extant SISR models. The principal technical contributions of this research are outlined as follows:
\begin{itemize}
\item We introduce a set of straightforward, yet effective self-supervised constraints designed to augment existing SR models. These constraints specifically target and refine areas of uncertainty encountered during the training process.
\item We expand the conventional SISR framework to incorporate a dual asymmetric paradigm. This paradigm comprises an online SR network complete with a projection head and a target SR network. The latter is progressively updated from the online network using the EMA strategy, thereby generating pseudo-targets that underpin the self-supervised constraints.
\item Our proposed paradigm is modular and easily integrated, making it compatible with any established SR model. Empirical results demonstrate that representative SR networks, when retrained with our Self-Supervised Constraint (SSC) paradigm, consistently achieve measurable improvements across all evaluated datasets.
\end{itemize}

\section{Related Work\label{sec:related_work}}

\noindent\textbf{Single Image Super-Resolution}
SISR is a fundamental research topic, and deep-learning-based methods have involved dramatic improvements and dominated the SISR in recent years \cite{survey_sisr_deep21}. Dong \textit{et al.} \cite{SRCNN} proposed pioneering work in super-resolution by introducing a fully convolutional network composed of three convolutional layers, named SRCNN \cite{SRCNN}. Subsequently, more effective SR models have been studied \cite{EDSR,RCAN,SAN,HAN,DBLP:journals/corr/abs-2307-16140,NLSN,8936424,9229100,9515582}. Most recently, Transformer-Based architectures have attracted great attention and made great breakthroughs in image super-resolution tasks \cite{SwinIR,HAT}. In this paper, we do not pay special attention to architectures, but explore an explicit self-supervised constraint to enhance existing SR methods.

\begin{figure*}
    \centering
    \includegraphics[width=0.825\textwidth]{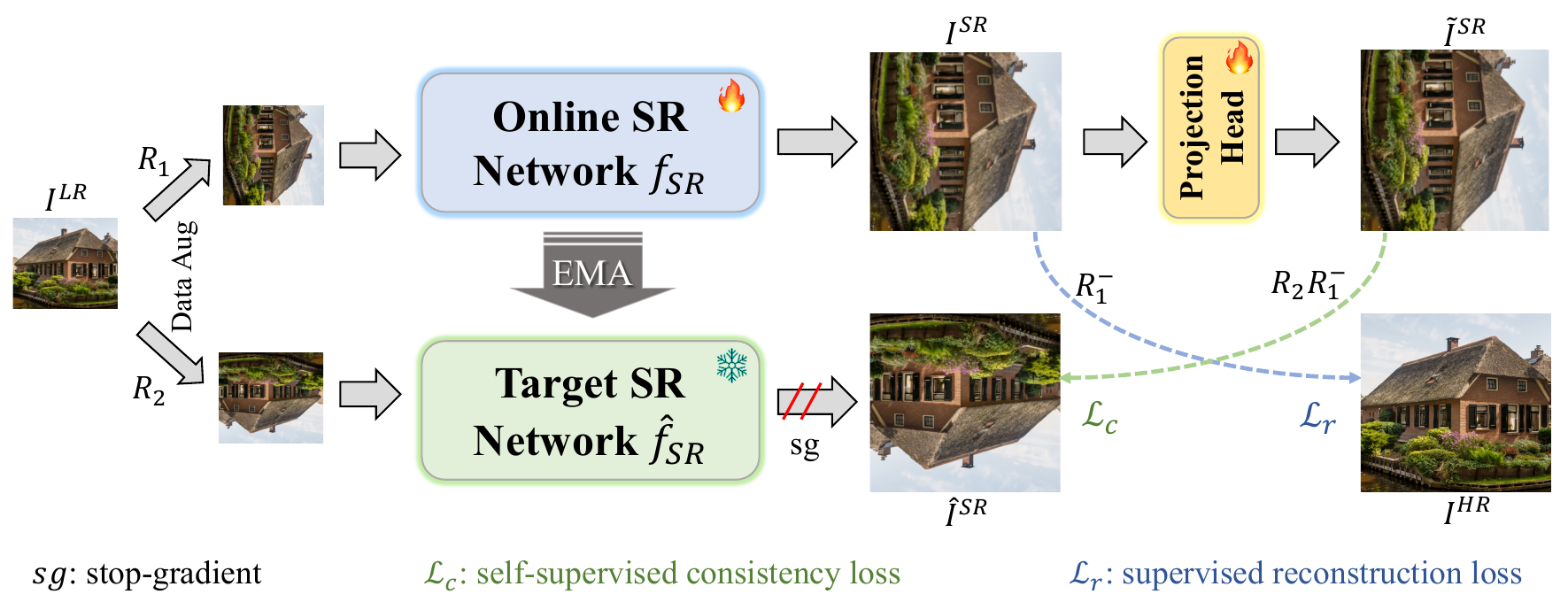}
    \vspace{-2mm}
    
    \caption{Overview of our proposed method SSC-SR. We adopt the same training framework from the BYOL where there are an online SR network $f_{SR}$, a target SR network $\hat{f}_{SR}$ and an asymmetric projection head $f_{proj}$ are adopted. The target SR network $\hat{f}_{SR}$ is updated via the exponential moving average (EMA) strategy. The online SR network $f_{SR}$ is trained with a pixel-wise loss $\mathcal{L}_{p}$ between the super-resolved image $I^{SR}$ and the ground truth image $I^{HR}$ and an additional consistency loss $\mathcal{L}_{c}$ calculated by the projected image $\tilde{I}^{SR}$ and target image $\hat{I}^{SR}$.}
    \label{fig:framework}
    \vspace{-2mm}
\end{figure*}
\noindent\textbf{Self-Supervised Learning} Recent advancements in self-supervised learning, a domain that leverages pretext tasks instead of manual annotations, have gained significant attention \cite{contrastive_survey}. Contrastive learning, especially, has emerged as a powerful approach, achieving remarkable success in various visual tasks, including both high-level and low-level tasks \cite{MoCo,SimCLR,Zero-Mean,contrastive_dehazing,contrastive_dehazing_cvpr23,PCL,MCIR,Contrastive_Derain_ICLR24}. These methods typically involve distinguishing positive samples from numerous negatives to prevent trivial solutions. In the field of image restoration, Wu \emph{et al.} \cite{contrastive_dehazing} applied contrastive learning to image dehazing, using ground truth images as positive samples and hazy inputs as negatives. Wu \textit{et al.} introduced the contrastive learning framework for single image super-resolution by hard negative mining strategy \cite{PCL}. They proposed the model contrastive for image restoration by revisiting the role of latency models \cite{MCIR}. These approaches generally emphasize negative sample mining to enrich the learning process. In this paper, our SSC-SR method diverges from this trend. Instead of relying on negative samples, we focus on harnessing uncertainty in the SR learning process. SSC-SR introduces a simple yet effective self-supervised constraint by providing suitable pseudo positives, effectively addressing the ill-posed nature of super-resolution tasks.

\section{Method\label{sec:related_method}}

\subsection{Overall Architecture}

We adopt the asymmetric dual architecture which contains the online SR network $f_{SR}$ followed by a projection head $f_{proj}$ and a target SR network $\hat{f}_{SR}$ which is the same architecture as the online model but updated with EMA strategy. The main framework of our proposed SSC-SR method is shown in \cref{fig:framework}. Given the training image $L^{LR}$, we first produce two augmented samples by applying rotation augmentation. The first sample is processed by the online SR network to produce a super-resolved image, which is then passed through the projection head. This projection head, unique to the online SR branch, creates an asymmetric dynamic with the target network. The target SR network processes the second augmented sample, resulting in another super-resolved image. Training the entire SR network involves a self-supervised consistency loss, comparing the online network's projected result with the target network's output, and a supervised reconstruction loss, comparing the online network's output with the ground truth.  Finally, the entire SR network is trained with self-supervised driven consistency loss (between the projected result of the online SR network and the output of the target SR network) and supervised reconstruction loss (between the output of the online SR network and the ground truth).

In our model, the online SR network ($f_{SR}$) and the target network ($\hat{f}_{SR}$) are defined by the parameters $\theta$ and $\hat{\theta}$, respectively. The target network, updated using an exponential moving average (EMA) strategy, is instrumental in training the online network by providing regression targets. This update is mathematically represented as
\begin{equation}
    \hat{\theta}^{t} = \beta\hat{\theta}^{t-1}+(1-\beta)\theta^{t-1},
\end{equation}
where $t$ denotes the training step and $\beta \in [0, 1]$ is the EMA decay rate. Notably, there is an projection head $f_{proj}$ following and updating with the online network simultaneously. The projection head $f_{proj}$ is simply stacked with three convolutional layers.

\subsection{Data Augmentation}
\label{sec:dataAug}

Our method employs explicit image augmentation to achieve self-supervised constraint by random flip and rotation operation, denoted as function $\mathrm{R}$. For a given low-resolution training image $I^{LR}$, we generate two variations, $\mathrm{R_{1}}(I^{LR})$ and $\mathrm{R_{2}}(L^{LR})$, as inputs for the online and target branches. This results in two super-resolved images, $I^{SR}$ and $\hat{I}^{SR}$, enhancing the training process by providing varied perspectives for model learning. In detail, super-resolved results are obtained as follows:
\begin{equation}
\begin{aligned}
     I^{SR} =&~f_{SR}(\mathrm{R}_{1}(I^{LR})),\\ 
     \hat{I}^{SR}=&~\hat{f}_{SR}(\mathrm{R}_{2}(I^{LR})),
\end{aligned}
\end{equation}
where $f^{SR}$ and $\hat f^{SR}$ denote the online and target SR networks, respectively.

For the online branch, an asymmetric predictor is applied to the output of the online SR network to obtain a prediction $\tilde{I}^{SR}$ of $\hat{I}^{SR}$, 
\begin{equation}
\begin{aligned}
     \tilde{I}^{SR}=&~f_{proj}(I^{SR}).\\
\end{aligned}
\end{equation}

\subsection{Self-Supervised Constraint}
The essential loss function of our SSC-SR framework is the reconstruction loss between the output and the ground truth, which is also the loss of the backbone online SR network. In addition to the reconstruction loss, the proposed SSC-SR additionally introduces a self-supervised driven consistency loss to improve the training of the backbone online SR network. Our intuition is to explicitly make the outputs of original and augmented samples the same in the training of one batch. 

It is worth noting that this differs from the commonly used ensemble learning strategies in existing SR methods, and they only add the augmented samples to the training dataset. Our framework extends existing SR models to further exploit the gains brought by data augmentation by explicitly introducing self-supervised constraints. At the same time, as a fundamental paradigm, our proposed method can further explore and utilize more augmentation strategies suitable for low-level image processing tasks.

Specifically, to ensure consistency between the rotated super-resolved image and the ground truth image, we reverse super-resolved images before calculating losses. Here, we note $R^{-}$ with a superscript as the reversed augmentation function. Therefore, the supervised reconstruction loss is calculated between the super-resolved image $I^{SR}$ and the ground truth image $I^{HR}$ with $L_{1}$ loss:
\begin{equation}
    \mathcal{L}_{r}=L_{1}(\mathrm{R}_{1}^{-}(I^{SR}),~I^{HR}).
\end{equation}
Then our self-supervised driven consistency loss is calculated with $\tilde{I}^{SR}$ and $\hat{I}^{SR}$ as follows:
\begin{equation}
    \mathcal{L}_{c}=L_{1}(\mathrm{R}_{2}(\mathrm{R}_{1}^{-}(\tilde{I}^{SR})),~\hat{I}^{SR}).
\end{equation}
Finally, the total loss is:
\begin{equation}
    \mathcal{L} = \mathcal{L}_{r}+\alpha\mathcal{L}_{c},
\end{equation}
where the $\alpha$ is the balancing weight, and we use 0.01 as default. When $\alpha =0$, the proposed SSC-SR will reduce to the backbone online SR method.

\subsection{Some Remarks}
In this study, we revisit the learning processes of SR models, enhancing the vanilla end-to-end framework with an innovative asymmetric dual architecture. This approach, exploiting self-supervised constraints, significantly improves SISR models. In particular, our SSC-SR is versatile and can be applied to various existing SR models. We benchmark against advanced SISR models, retraining them within our SSC-SR framework to validate its effectiveness. In addition, we do not pay special attention to developing optimal data augmentation strategies. Instead, we employ commonly used rotation augmentation for a fair and consistent comparison across existing methods.

\begin{table*}[ht]
\centering
\caption{Quantitative comparison with existing methods on five widely used benchmark datasets. Some advanced SR approaches are retrained and enhanced by our proposed SSC-SR method. Results of our retrained models are in \textbf{bold}.} 
\label{tab:main_results}
\resizebox{\textwidth}{!}{
\begin{tabular}{|c|c|c|c|c|c|c|c|c|c|c|c|c|c|}
\hline
\multirow{2}{*}{Method} & \multirow{2}{*}{Dataset}  & \multicolumn{2}{c|}{Set5} & \multicolumn{2}{c|}{Set14} & \multicolumn{2}{c|}{B100} & \multicolumn{2}{c|}{Urban100} & \multicolumn{2}{c|}{Manga109}  & \multicolumn{2}{c|}{Average}  \\
\cline{3-14}
   & & PSNR  & SSIM             & PSNR  & SSIM              & PSNR  & SSIM             & PSNR  & SSIM                 & PSNR  & SSIM     & PSNR  & SSIM               \\
\hline
\hline
SAN \cite{SAN} & DIV2K   & 32.64 & 0.9003        & 28.92 & 0.7888            & 27.78 & 0.7436           & 26.79 & 0.8068               & 31.18 & 0.9169           & 29.462	&0.8313
   \\
HAN \cite{HAN} & DIV2K & 32.64 & 0.9002           & 28.90 & 0.7890            & 27.80 & 0.7442           & 26.85 & 0.8094             & 31.42 & 0.9177       &29.522	&0.8321
    \\
IGNN \cite{IGNN} & DIV2K
& {32.57}
& {0.8998}
& {28.85}
& {0.7891}
& {27.77}
& {0.7434}
& {26.84}
& {0.8090}
& {31.28}
& {0.9182} & 29.462 &	0.8319      \\
\hline
EDSR \cite{EDSR} &\multirow{3}{*}{DIV2K}  & 32.46 & 0.8968         & 28.80 & 0.7876            & 27.71 & 0.7420           & 26.64 & 0.8033               & 31.02 & 0.9148  & 29.33	&0.8289           \\
 \textbf{EDSR (Ours)} & & \textbf{32.56} & \textbf{0.8986 }          & \textbf{28.87} & \textbf{0.7889}           & \textbf{27.75} &\textbf{ 0.7437}           & \textbf{26.75} &\textbf{0.8070}              & \textbf{31.22} & \textbf{0.9173}      &   \textbf{29.43}	& \textbf{0.8311}       \\
\emph{Improvements}   &  & \textcolor{blue}{0.10}  & \textcolor{blue}{0.0018}          & \textcolor{blue}{0.07}  & \textcolor{blue}{0.0013}            & \textcolor{blue}{0.04}  & \textcolor{blue}{0.0017}           & \textcolor{blue}{0.11}  &\textcolor{blue}{0.0037}               & \textcolor{blue}{0.20}  & \textcolor{blue}{0.0025} &  \textcolor{blue}{0.10}	 & \textcolor{blue}{0.0022} \\

\hline
\hline

RCAN \cite{RCAN}&\multirow{3}{*}{DIV2K} & 32.63 & 0.9002           & 28.87 & 0.7889            & 27.77 & 0.7436           & 26.82 & 0.8087               & 31.22 & 0.9173            &     29.46	        & 0.8317        \\
\textbf{RCAN (Ours)} &  & \textbf{32.66} & \textbf{0.9005} & \textbf{28.91} & \textbf{0.7894}            & \textbf{27.78} & \textbf{0.7445}          & \textbf{26.88} &\textbf{0.8094}             & \textbf{31.38} & \textbf{0.9187}     & \textbf{29.52}	    & \textbf{0.8325} \\
\emph{Improvements}  &   & \textcolor{blue}{0.03}  & \textcolor{blue}{0.0003}           & \textcolor{blue}{0.04}  & \textcolor{blue}{0.0005}            & \textcolor{blue}{0.01}  & \textcolor{blue}{0.0009}           & \textcolor{blue}{0.06}  & \textcolor{blue}{0.0007}               & \textcolor{blue}{0.16}  & \textcolor{blue}{0.0014}       &    \textcolor{blue}{0.06}	  &\textcolor{blue}{0.0008}        \\

\hline
\hline

NLSN \cite{NLSN} &\multirow{3}{*}{DIV2K}  & 32.59 & 0.9000         & 28.87 & 0.7891            & 27.78 & 0.7444           & 26.96 & 0.8109               & 31.27 & 0.9184  &  29.49&	0.8326         \\

 \textbf{NLSN (Ours)}&  & \textbf{32.70} & \textbf{0.9009}         & \textbf{28.98} & \textbf{0.7911}           & \textbf{27.83} & \textbf{0.7466}           & \textbf{27.11} & \textbf{0.8153}             & \textbf{31.56} & \textbf{0.9216}   &  29.63 & 0.8351       \\
\emph{Improvements} &    & \textcolor{blue}{0.11}  & \textcolor{blue}{0.0009}          & \textcolor{blue}{0.11}  & \textcolor{blue}{0.0020}            & \textcolor{blue}{0.05}  & \textcolor{blue}{0.0022}           & \textcolor{blue}{0.15}  &\textcolor{blue}{0.0044}               & \textcolor{blue}{0.29}  & \textcolor{blue}{0.0032} & \textcolor{blue}{0.14} &	\textcolor{blue}{0.0025}\\

\hline
\hline

SwinIR \cite{SwinIR} &\multirow{3}{*}{DIV2K} 
& 32.72
& 0.9021
& 28.94
& 0.7914
& 27.83
& 0.7459
& 27.07
& 0.8164
& 31.67
& 0.9226 
& 29.65
& 0.8357
\\

 \textbf{SwinIR (Ours)} &  & \textbf{32.79} & \textbf{0.9025}         & \textbf{29.00} & \textbf{0.7921}           & \textbf{27.84} & \textbf{0.7463}           & \textbf{27.12} & \textbf{0.8173}             & \textbf{31.74} & \textbf{0.9233}    & \textbf{29.71} & \textbf{0.8363}          \\
\emph{Improvements}     & 
& \textcolor{blue}{0.07}  & \textcolor{blue}{0.0004}          & \textcolor{blue}{0.06}  & \textcolor{blue}{0.0007}            & \textcolor{blue}{0.01}  & \textcolor{blue}{0.0004}           & \textcolor{blue}{0.06}  &\textcolor{blue}{0.0010}               & \textcolor{blue}{0.08}  & \textcolor{blue}{0.0007}&
\textcolor{blue}{0.06}	& \textcolor{blue}{0.0006} \\

\hline
\hline

HAT \cite{HAT} & \multirow{3}{*}{DF2K} &
32.92 & 0.9047  & 29.15  & 0.7958  & 27.97 &  0.7505  & 27.87  & 0.8346 &  32.35 &  0.9283 & 30.05	& 0.8428 \\

 \textbf{HAT (Ours)}&   & \textbf{32.95} & \textbf{0.9049}         & \textbf{29.18} & \textbf{0.7961}           & \textbf{27.98} & \textbf{0.7508}           & \textbf{27.91} & \textbf{0.8351}             & \textbf{32.37} & \textbf{0.9285}  &  \textbf{30.08}   & \textbf{0.8431}            \\
\emph{Improvements}  &  
& \textcolor{blue}{0.03}  & \textcolor{blue}{0.0002}          & \textcolor{blue}{0.03}  & \textcolor{blue}{0.0003}            & \textcolor{blue}{0.01}  & \textcolor{blue}{0.0003}           & \textcolor{blue}{0.04}  &\textcolor{blue}{0.0005}               & \textcolor{blue}{0.02}  & \textcolor{blue}{0.0002}&
\textcolor{blue}{0.03}	& \textcolor{blue}{0.0003}\\

\hline
\end{tabular}
}
\end{table*}

\section{Experiments\label{sec:related_experiments}}
In this section we describe the detailed evaluation experiments. Firstly, we introduce the experiment settings and comparison methods. Then quantitative and qualitative results are reported. Lastly, ablation studies about the impact of different components are summarized.
\subsection{Experiment Setup}

\textbf{Training Detail} Following comparison methods \cite{EDSR,SwinIR}, we crop HR patches with a fixed size of $48\times~48$ for training. We set the batch size to 16 and train it using ADAM optimizer with the settings of $\beta_1$ = 0.9, $\beta_2$ = 0.999. The initial learning rate is set as $10^{-4}$. For comparison, we measured the peak signal-to-noise ratio (PSNR) and the structural similarity index measure (SSIM) on the Y channel of transformed YCbCr space. 

\noindent \textbf{Comparison Method} Since our proposed SSC-SR is generic and can be applied to any existing method. We retrain EDSR~\cite{EDSR} and RCAN \cite{RCAN}, NLSN \cite{NLSN}, SwinIR \cite{SwinIR}, and HAT\cite{HAT} by our proposed SSC-SR framework. In addition, we also introduce some representative work for comparisons, including SAN \cite{SAN}, HAN \cite{HAN}, IGNN \cite{IGNN}.

\subsection{Main Results}
In our study, we present the results of retrained models, specifically EDSR \cite{EDSR}, RCAN \cite{RCAN}, NLSN \cite{NLSN}, SwinIR \cite{SwinIR}, and HAT \cite{HAT}, as illustrated in \cref{tab:main_results}. These models were tested in five benchmark datasets. Upon comparison with existing methods, it is evident that our retrained models exhibit superior performance in terms of PSNR and SSIM across all test datasets. This underscores the effectiveness of the proposed self-supervised constraint in improving model performance. Specifically, we conducted an evaluative study of the SSC-SR with respect to different baseline architectures and datasets. The results show that CNN-based methods, including EDSR, RCAN, and NLSN, demonstrate an average gain of nearly 0.1 dB in PSNR. A similar level of improvement is observed in the Transformer-based models, SwinIR and HAT, indicating the broad applicability and efficiency of our approach.

Furthermore, our comparative analysis of the reconstruction results on specific datasets, notably Manga109 and Urban100, reveals significant improvements achieved through our SSC-SR. Specifically, the EDSR, RCAN, and NLSN models exhibit remarkable gains of 0.20 dB, 0.16 dB, and 0.29 dB in the Manga109 dataset, respectively. Moreover, a consistent and notable improvement is observed across all retrained models on the Urban100 dataset. This enhancement can be attributed to the dataset characteristics, such as the prevalence of edges and textures in Urban100. Our self-supervised constraint, which is data augmentation-based, effectively leverages these features during training. This strategy not only boosts the model's performance but also serves as a 'free lunch' for enhancing existing methods. In addition, our comparative analysis showcases the superior performance of our retrained RCAN model against newer methods such as SAN \cite{SAN}. Notably, the retrained RCAN outperforms these newer models, while the retrained NLSN surpasses the Transformer-based SwinIR in all test datasets except Set5. demonstrating the practicality and efficacy of our approach.

Interestingly, the overall improvement in Transformer-based methods, including HAT, is less pronounced compared to CNN-based models. This observation can be explained from two perspectives. Firstly, Transformer-based architectures inherently capture longer-range relationships more effectively than CNN-based models, complementing our self-supervised constraint and yielding high-fidelity results. Secondly, the use of the extensive DF2K dataset in training HAT likely brings it closer to an optimal solution. However, our self-supervised constraint facilitates an improvement of the DIV2K dataset without the need for additional data. These results underline the effectiveness of our proposed SSC-SR framework. It harnesses a self-supervised driven consistency loss to enhance existing models, demonstrating its applicability and advantage in the field of image restoration.

\begin{figure*}[!t]
\centering
\includegraphics[width=0.975\textwidth]{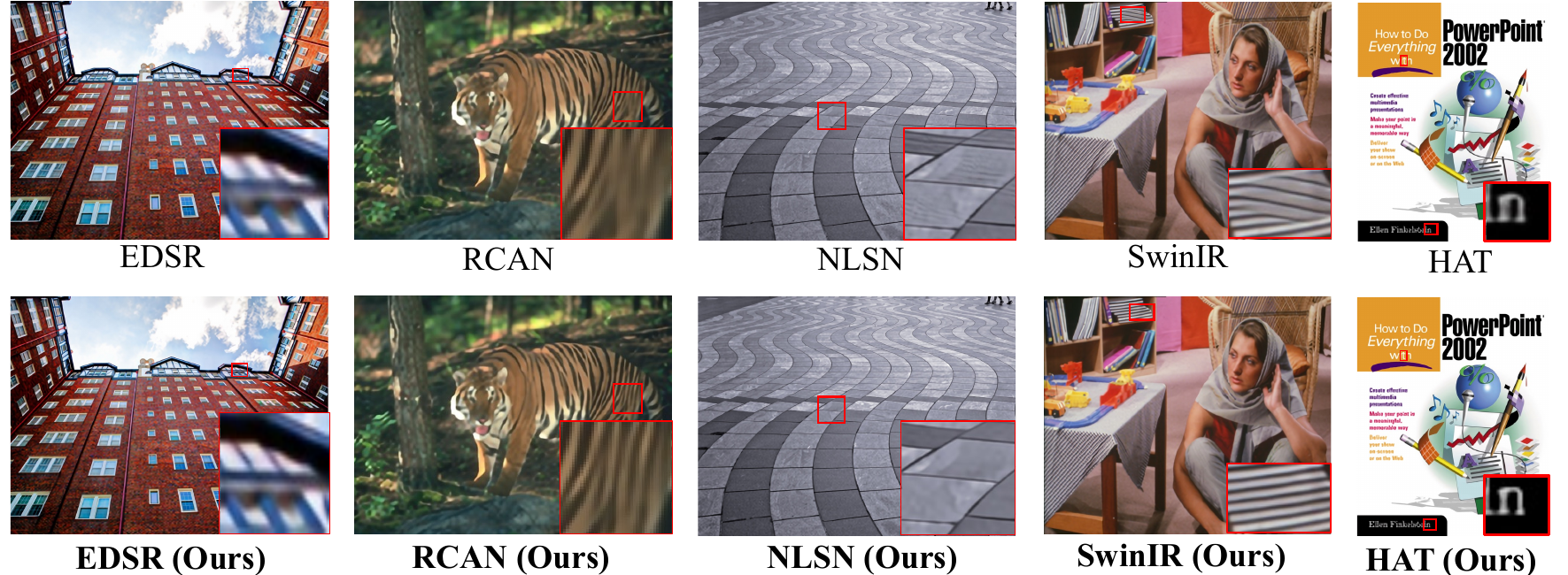}
    \vspace{-2mm}

 \caption{Visual comparisons between our enhanced models and their original counterparts are presented. The first row displays the results of existing methods, while the second row showcases the corresponding improvements achieved by our retrained models (Zoom in for more detail).}\label{fig:visual_results}
 
\end{figure*}

\begin{table}
\centering
\vspace{-2mm}

\caption{Qualitative comparison in term of LPIPS between benchmark methods and our retrained counterparts. }\label{tab:lpips}
\vspace{-2mm}
\begin{center}

\resizebox{0.4\textwidth}{!}
{
\begin{tabular}{|cccc|}
\hline 
Model     & B100 & Urban100 & Manga109 \\
\hline
\hline
EDSR        & 0.3022 &  0.2356  & 0.1364    \\
\textbf{EDSR (Ours)}   & 0.3019 &  0.2354  & 0.1360    \\
\hline

RCAN        & 0.3022 &  0.2324  & 0.1353    \\
\textbf{RCAN (Ours)}    & 0.3009 &  0.2306  & 0.1322    \\
\hline

NLSN        & 0.2969 &  0.2261  & 0.1309   \\
\textbf{NLSN (Ours)}    & 0.2966 &  0.2247  & 0.1300 \\
\hline

\hline
\end{tabular}
}
\end{center}
    \vspace{-6mm}

\end{table}

\begin{table}\normalsize
\centering
\caption{Effectiveness of the EMA strategy.
\label{tab:ema_cof}}
\begin{center}
\begin{tabular}{|cccc|}
\hline
Updating & PSNR $\uparrow$ & SSIM$\uparrow$ & LPIPS $\downarrow$  \\
\hline
\hline

w/o EMA ($\beta$=0) & 32.53 & 0.8983 & 0.1801 \\
w EMA ($\beta$=0.999) & \textbf{32.56} & \textbf{0.8986} & \textbf{0.1799}\\
\hline
\end{tabular}
\end{center}
\end{table}

\begin{figure}[!t]
    \centering
    \includegraphics[width=0.475\textwidth]{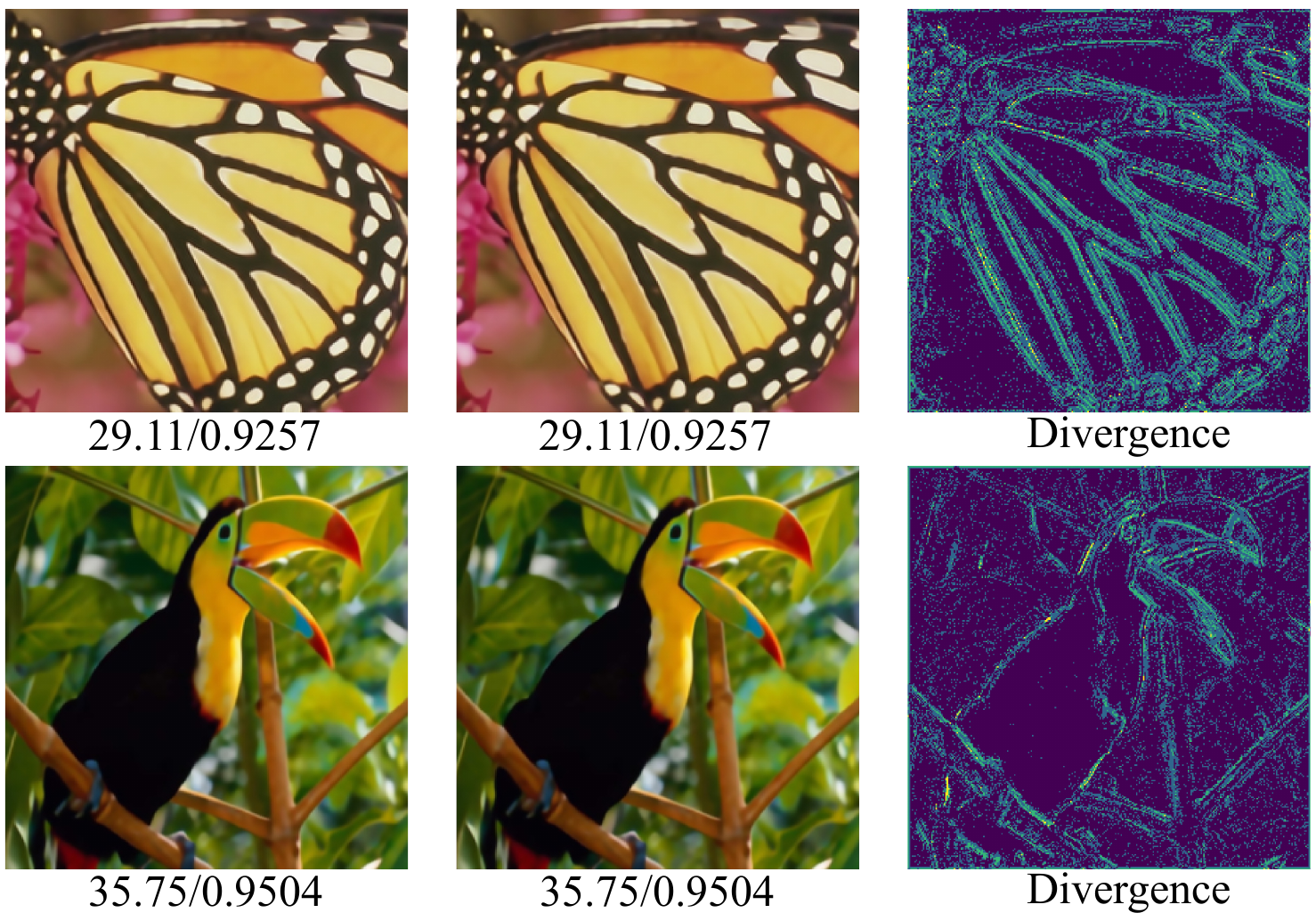}
    \caption{The figure presents a side-by-side visual comparison of the super-resolved images from an online SR network and their counterparts generated by the projection head, concluding a divergence row with a divergence map for comparison.}
    \label{fig:projection_img}
\end{figure}

\begin{table}[t]\normalsize
\centering
\caption{The influence of measurement on self-supervised driven consistency loss.
}\label{tab:measure}

\begin{tabular}{|ccccc|}
 \hline
 Metric & $L_1$ & $L_2$ & $\mathrm{VGG}-L_{1}$ & $\mathrm{VGG}-L_{2}$   \\
\hline
\hline
PSNR & \textbf{ 32.56} & - & 32.54 & 32.53  \\ 
SSIM & \textbf{0.8986} & - & \textbf{0.8986} & 0.8984  \\ 
LPIPS & 0.1799 & - & 0.1795 & \textbf{0.1790}  \\ 
\hline

\end{tabular}
\vspace{-2mm}
\end{table}

To demonstrate the efficacy of our method, we present visual comparisons between benchmark models and our retrained versions. Illustrated in \cref{fig:visual_results}, our retrained models, benefiting from self-supervised constraints and data augmentation, show notably clearer and sharper textures. Particularly in cases where original methods falter in edge definition, our retrained models excel at restoring precise textures. For instance, in the first row of \cref{fig:visual_results}, our retrained EDSR model correctly maintains edge details where the original version fails. Similarly, other retrained models exhibit enhanced texture details and fewer artifacts compared to their original counterparts.

The results, as detailed in \cref{tab:main_results}, demonstrate that our SSC-SR method significantly enhances the performance of basic models, particularly evidenced on the Urban100 test dataset. Additional visual comparisons in \cref{fig:visual_results} reveal that our retrained models yield clearer, more accurate textures with fewer artifacts. Furthermore, the LPIPS values for EDSR, RCAN, NLSN, and their retrained counterparts, presented in \cref{tab:lpips}, confirm superior qualitative performance. This improvement is attributed to our self-supervised constraints focusing on complex areas, typically textures. Thus, one can conclude that SSC-SR boosts both quantitative and qualitative aspects of basic models.

\subsection{Ablation Study}

\begin{table}[t]
\centering
\caption{Different implementations of projection head $f_{proj}$.}\label{tab:projection}
\begin{tabular}{|ccc|}
\hline
Metric & Conv & DConv    \\
\hline
\hline 
PSNR & 32.52 & \textbf{32.56 }  \\ 
SSIM  & 0.8983 & \textbf{0.8986 }  \\ 
LPIPS  & \textbf{0.1796} & 0.1799  \\ 
\hline
\end{tabular}
\vspace{-2mm}

\end{table}

\noindent\textbf{Effectiveness of the EMA.}
In our SSC-SR, target model is updated from online model with EMA. Here we evaluate different updating coefficients of EMA and detailed results are reported in \cref{tab:ema_cof}. When $\beta=0$, the proposed method reduces to the case where the online and the target networks share the same parameters. From the comparison results, we can learn that the EMA strategy can improve performance.

\noindent\textbf{Different Measurements.}
To assess the effectiveness of our self-supervised consistency loss within the SSC-SR framework, we conducted experiments by retraining the EDSR model, particularly focusing on loss function variations. Our analysis revealed that the loss $L_{1}$ led to the most superior performance. Conversely, the use of $L_{2}$ loss resulted in considerably lower PSNR scores. Incorporating a pre-trained VGG network for loss calculation in latent space, designated as $\mathrm{VGG}-L_{1}$ and $\mathrm{VGG}-L_{2}$, yielded marginal quantitative improvements but noticeably better qualitative results. This underscores the efficacy of self-supervised constraints, especially in intricate regions, by using pseudo HR images generated from the target EMA model.

\noindent\textbf{Projection Head.}
Our ablation analysis scrutinized the efficacy of different projection head implementations, specifically comparing convolution (Conv) and deformable convolution (DConv) layers. The investigation reveals that DConv outperforms Conv, as shown in \cref{tab:projection}. This indicates the projection head's crucial role in addressing the 'one-to-many' mapping challenge, by offering multiple valid high-resolution targets to the online SR network. Although PSNR/SSIM metrics suggest similarity between super-resolved outcomes and projection head outputs, their distinct details, highlighted in normalized error maps, confirm their separate solution validity. Future studies will delve into the impact of the projection head's architecture on overall performance.

\section{Conclusion\label{sec:related_conclusion}}
In this paper, we present SSC-SR, an innovative self-supervised constraint for super-resolution, harnessing data augmentation and training uncertainties. This method, grounded in self-supervised consistency loss driven by data augmentation, has proven to be simple yet highly effective. Our thorough experiments validate the enhanced performance of retrained models using SSC-SR. Furthermore, detailed ablation studies corroborate its efficacy. Moving forward, our goal is to extend the application of our strategies to a broader range of state-of-the-art SISR methods and other low-level image restoration tasks.

\section{ACKNOWLEDGMENT}
The research was supported by the National Natural Science Foundation of China (U23B2009, 92270116).

{
\scriptsize
\bibliographystyle{IEEEbib}
\bibliography{sisr}

\begin{thebibliography}{10}

\bibitem{survey_sisr_deep21}
Juncheng Li, Zehua Pei, and Tieyong Zeng,
\newblock ``From beginner to master: {A} survey for deep learning-based single-image super-resolution,''
\newblock {\em CoRR}, vol. abs/2109.14335, 2021.

\bibitem{ACMComputingSurvey}
Saeed Anwar, Salman~H. Khan, and Nick Barnes,
\newblock ``A deep journey into super-resolution: {A} survey,''
\newblock {\em {ACM} Comput. Surv.}, 2020.

\bibitem{EDSR}
Bee Lim, Sanghyun Son, Heewon Kim, Seungjun Nah, and Kyoung~Mu Lee,
\newblock ``Enhanced deep residual networks for single image super-resolution,''
\newblock in {\em CPVR}, 2017.

\bibitem{RCAN}
Yulun Zhang, Kunpeng Li, Kai Li, Lichen Wang, Bineng Zhong, and Yun Fu,
\newblock ``Image super-resolution using very deep residual channel attention networks,''
\newblock in {\em {ECCV}}, 2018.

\bibitem{NLSN}
Yiqun Mei, Yuchen Fan, and Yuqian Zhou,
\newblock ``Image super-resolution with non-local sparse attention,''
\newblock in {\em {CVPR}}, 2021.

\bibitem{SwinIR}
Jingyun Liang, Jiezhang Cao, Guolei Sun, Kai Zhang, Luc~Van Gool, and Radu Timofte,
\newblock ``{SwinIR}: Image restoration using swin transformer,''
\newblock in {\em {ICCVW}}, 2021.

\bibitem{HAT}
Xiangyu Chen, Xintao Wang, Jiantao Zhou, Yu~Qiao, and Chao Dong,
\newblock ``Activating more pixels in image super-resolution transformer,''
\newblock in {\em {CVPR}}, 2023.

\bibitem{contrastive_survey}
Jie Gui, Tuo Chen, Qiong Cao, Zhenan Sun, Hao Luo, and Dacheng Tao,
\newblock ``A survey of self-supervised learning from multiple perspectives: Algorithms, theory, applications and future trends,''
\newblock {\em CoRR}, vol. abs/2301.05712, 2023.

\bibitem{contrastive_dehazing}
Haiyan Wu, Yanyun Qu, Shaohui Lin, Jian Zhou, Ruizhi Qiao, Zhizhong Zhang, Yuan Xie, and Lizhuang Ma,
\newblock ``Contrastive learning for compact single image dehazing,''
\newblock in {\em {CVPR}}, 2021.

\bibitem{PCL}
Gang Wu, Junjun Jiang, and Xianming Liu,
\newblock ``A practical contrastive learning framework for single-image super-resolution,''
\newblock {\em IEEE TNNLS}, pp. 1--12, 2023.

\bibitem{MCIR}
Gang Wu, Junjun Jiang, Kui Jiang, and Xianming Liu,
\newblock ``Learning from history: Task-agnostic model contrastive learning for image restoration,''
\newblock in {\em AAAI}, 2024.

\bibitem{SRCNN}
Chao Dong, Chen~Change Loy, Kaiming He, and Xiaoou Tang,
\newblock ``Image super-resolution using deep convolutional networks,''
\newblock {\em {IEEE} TPAMI}, vol. 38, no. 2, pp. 295--307, 2016.

\bibitem{SAN}
Tao Dai, Jianrui Cai, Yongbing Zhang, Shu{-}Tao Xia, and Lei Zhang,
\newblock ``Second-order attention network for single image super-resolution,''
\newblock in {\em {CVPR}}, 2019.

\bibitem{HAN}
Ben Niu, Weilei Wen, Wenqi Ren, Xiangde Zhang, Lianping Yang, Shuzhen Wang, Kaihao Zhang, Xiaochun Cao, and Haifeng Shen,
\newblock ``Single image super-resolution via a holistic attention network,''
\newblock in {\em {ECCV}}, 2020.

\bibitem{DBLP:journals/corr/abs-2307-16140}
Gang Wu, Junjun Jiang, Kui Jiang, and Xianming Liu,
\newblock ``Fully 1{\texttimes}1 convolutional network for lightweight image super-resolution,''
\newblock {\em CoRR}, vol. abs/2307.16140, 2023.

\bibitem{8936424}
Kui Jiang, Zhongyuan Wang, Peng Yi, Guangcheng Wang, Ke~Gu, and Junjun Jiang,
\newblock ``Atmfn: Adaptive-threshold-based multi-model fusion network for compressed face hallucination,''
\newblock {\em IEEE TMM}, vol. 22, no. 10, pp. 2734--2747, 2020.

\bibitem{9229100}
Kui Jiang, Zhongyuan Wang, Peng Yi, Tao Lu, Junjun Jiang, and Zixiang Xiong,
\newblock ``Dual-path deep fusion network for face image hallucination,''
\newblock {\em IEEE TNNLS}, vol. 33, no. 1, pp. 378--391, 2022.

\bibitem{9515582}
Kui Jiang, Zhongyuan Wang, Peng Yi, Chen Chen, Zheng Wang, Xiao Wang, Junjun Jiang, and Chia-Wen Lin,
\newblock ``Rain-free and residue hand-in-hand: A progressive coupled network for real-time image deraining,''
\newblock {\em IEEE TIP}, vol. 30, pp. 7404--7418, 2021.

\bibitem{MoCo}
Kaiming He, Haoqi Fan, Yuxin Wu, Saining Xie, and Ross~B. Girshick,
\newblock ``Momentum contrast for unsupervised visual representation learning,''
\newblock in {\em {CVPR}}, 2020.

\bibitem{SimCLR}
Ting Chen, Simon Kornblith, Mohammad Norouzi, and Geoffrey~E. Hinton,
\newblock ``A simple framework for contrastive learning of visual representations,''
\newblock in {\em {ICML}}, 2020.

\bibitem{Zero-Mean}
Xiong Zhou, Xianming Liu, Feilong Zhang, Gang Wu, Deming Zhai, Junjun Jiang, and Xiangyang Ji,
\newblock ``Zero-mean regularized spectral contrastive learning: Implicitly mitigating wrong connections in positive-pair graphs,''
\newblock in {\em ICLR}, 2024.

\bibitem{contrastive_dehazing_cvpr23}
Yu~Zheng, Jiahui Zhan, Shengfeng He, Junyu Dong, and Yong Du,
\newblock ``Curricular contrastive regularization for physics-aware single image dehazing,''
\newblock in {\em {CVPR}}. 2023, pp. 5785--5794, {IEEE}.

\bibitem{Contrastive_Derain_ICLR24}
Wu~Ran, Peirong Ma, Zhiquan He, Hao Ren, and Hong Lu,
\newblock ``Harnessing joint rain-/detail-aware representations to eliminate intricate rains,''
\newblock in {\em ICLR}, 2024.

\bibitem{IGNN}
Shangchen Zhou, Jiawei Zhang, Wangmeng Zuo, and Chen~Change Loy,
\newblock ``Cross-scale internal graph neural network for image super-resolution,''
\newblock in {\em NeurIPS}, 2020.

\end{thebibliography}
}
\end{document}